\documentclass[conference]{IEEEtran}
\IEEEoverridecommandlockouts
\usepackage{cite}
\usepackage{amsmath,amssymb,amsfonts}
\usepackage{algorithmic}
\usepackage{graphicx}
\usepackage{textcomp}
\usepackage{xcolor}
\usepackage{enumitem}
\usepackage{hyperref}
\usepackage[noabbrev]{cleveref}

\usepackage{makecell} 
\usepackage{array} 
\usepackage{tabularx}
\usepackage{comment}

\def\bng{\bngx}

\font\bngx=bang10

\def\*#1*#2{o\null{#2}{#1}}

\def\sh#1{\setbox0=\hbox{#1}%
     \kern-.02em\copy0\kern-\wd0
     \kern.04em\copy0\kern-\wd0
     \kern-.02em\raise.0433em\box0 }

\def\BibTeX{{\rm B\kern-.05em{\sc i\kern-.025em b}\kern-.08em
    T\kern-.1667em\lower.7ex\hbox{E}\kern-.125emX}}
    
\begin{document}

\title{ CineXDrama: Relevance Detection and Sentiment Analysis of Bangla YouTube Comments on Movie-Drama using Transformers: Insights from Interpretability Tool}

\author{
\IEEEauthorblockN{Usafa Akther Rifa, Pronay Debnath, Busra Kamal Rafa, Shamaun Safa Hridi,
Md. Aminur Rahman}
Department of Computer Science and Engineering,\\
Ahsanullah University Of Science and Technology (AUST), Dhaka, Bangladesh\\
Email-  \{usafarifa97, pronaydebnath99, brafa263.3, shamaunsafahridi, aminur.rahman.rsd\}@gmail.com 
}
\maketitle

\begin{abstract}
In recent years, YouTube has become the leading platform for Bangla movies and dramas, where viewers express their opinions in comments that convey their sentiments about the content. However, not all comments are relevant for sentiment analysis, necessitating a filtering mechanism. We propose a system that first assesses the relevance of comments and then analyzes the sentiment of those deemed relevant. We introduce a dataset of 14,000 manually collected and preprocessed comments, annotated for relevance (relevant or irrelevant) and sentiment (positive or negative). Eight transformer models, including BanglaBERT, were used for classification tasks, with BanglaBERT achieving the highest accuracy (83.99\% for relevance detection and 93.3\% for sentiment analysis). The study also integrates LIME to interpret model decisions, enhancing transparency. 
\end{abstract}

\section{Introduction}
In today's world, YouTube has become a popular platform for entertainment, with many people using it to watch content. YouTube has more than 2.70 billion monthly active users as of October 2024, highlighting its massive global reach \cite{reach}. As Bangla movies and dramas have recently surged in popularity among both Bangladeshi people and Bangla-speaking communities worldwide, YouTube has become a key platform for streaming this content. Viewers frequently express their opinions in the comment section, sharing whether they enjoyed the drama or found it lacking. These comments represent a rich dataset in the Bangla language, offering valuable insights into viewers' sentiments and perceptions.\\
Analyzing YouTube comments in Bangla about Bangla dramas and movies helps reveal viewers' sentiments, identifying whether content is seen positively or negatively. An automated system using these comments could offer real-time feedback, helping filmmakers gauge audience reactions and influence the Bangladeshi entertainment industry with direct insights. Many works have already been done on sentiment analysis using YouTube comments \cite{ref2,ref3,ref4}, but there is limited research in the Bangla language, even though Bangla is the world’s sixth most spoken language, with over 237 million native speakers \cite{ref5}. Some work has been done on Bangla YouTube comments for sentiment analysis \cite{ref6,ref7,ref8}, but these efforts have not been specifically focused on both Bangla movies and dramas. Instead, they have covered a wide range of Bangla content, including dramas, movies, music, news, and more. As a result, these sentiment analysis projects do not specifically cater to the needs of Bangla movie or drama reviews.\\
Another key challenge is identifying relevant comments. This step is crucial for   accurately assessing audience feedback on whether a drama or movie is good or bad. The comments section often includes off-topic posts, inappropriate remarks, or religious or promotional content that doesn’t reflect the actual quality of the movie or drama. For example, comments like “{\bng 2022 sael ek ek edkhechn?}” (Who's watching in 2022?) or “{\bng Aaim musilm Hey gir/bt}” (I am proud to be Muslim) do not provide meaningful feedback on the content’s quality. Analyzing only the relevant comments is essential for a precise evaluation, as including irrelevant ones can skew the sentiment analysis and lead to an inaccurate understanding of the viewers true reactions. Ignoring off-topic or unrelated comments ensures the sentiment analysis remains focused on the actual feedback about the drama or movie. \\
In this study, we have created a large dataset consisting of 14,000 Bangla comments related to Bangla movies and dramas, which will be discussed in the Dataset section. Additionally, we have analyzed the relevance of the comments to determine whether they are pertinent to the content of the movies and dramas. This type of work has not been previously done in the Bangla language. For relevance and sentiment analysis, we utilized eight transformer models. These models offer the advantage of handling complex Bangla language structures, enabling more accurate sentiment classification and relevance detection. To improve transparency, we integrated LIME (Local Interpretable Model-Agnostic Explanations) \cite{Lime} as an interpretability tool. LIME helped explain how the models made predictions, increasing trust in the results. In summary, the key contributions of this paper are as follows:
\begin{itemize}
    \item We have created a dataset consisting of 14,000 Bangla YouTube comments from Bangla movies and dramas for the purposes of relevance detection and sentiment analysis. The dataset contains 7839 relevant and 6,161 irrelevant comments. Among the relevant comments, 4275 are labeled as positive, and 3564 are labeled as negative. 
    \item We have proposed a system that initially determines whether a comment is relevant or irrelevant using pre-trained transformer models. If the comment is identified as relevant, then classifies it into either a positive or negative category for sentiment analysis.
    \item Among the eight pre-trained transformers we utilized (mentioned in Subsection \ref{setup}), BanglaBERT performed the strongest in both relevance detection and sentiment analysis, with accuracies of 83.99\% and 93.3\% respectively.
    \item To enhance model interpretability, we used LIME for a visual understanding of predictions, enabling a clearer evaluation of feature importance and model behavior.
    
\end{itemize}

\section{Literature Review}
Numerous research studies have explored sentiment analysis and comment relevance detection across various languages and domains. In sentiment analysis, Nafis et al. \cite{Srel1} examined YouTube comments in English, Bangla, and Romanized Bangla using LSTM, achieving accuracies of 65.97\% for three sentiment labels and 54.24\% for five labels. Govindarajan et al. \cite{Srel3} enhanced sentiment analysis by combining the NB-GA method on 2,000 movie reviews, achieving a high accuracy of 93.80\%. Salim et al. \cite{Srel4} compared supervised classifiers on machine-translated English and authentic Bengali datasets, finding SVM to perform best with macro F1 scores of 0.923 and 0.916 and accuracy rates of 93.5\% for English and 93.0\% for Bengali. Rumman et al. \cite{Srel7} improved multilingual sentiment analysis by gathering around 4,000 social media samples, demonstrating that LSTM increased accuracy by 25\% compared to Nafis et al. \cite{Srel1}, underscoring the advantages of deep learning approaches. In the field of detecting relevant and non-relevant comments, Marzieh et al. \cite{Rrel1} introduced a framework incorporating syntactical, topical, and semantic features with word embeddings, achieving 86\% precision on 33,921 BBC News Facebook comments—an improvement of 9.6\% over previous methods. Their findings revealed that 60\% of comments were off-topic. Similarly, Marthe et al. \cite{Rrel2} applied supervised machine learning (SML) to assess the relevance of 182,367 YouTube comments on 305 music videos, achieving a relevance detection rate of 78\% to 83\% after manually annotating 40\% of the comments, highlighting the value of a broader relevance definition. Andrei et al. \cite{Rrel5} addressed the challenge of ranking YouTube comments by proposing a system that links comments to relevant web pages and compares three ranking methods. Their two-step approach, combining a neural network binary classifier with topic extraction and weighted relevance, significantly improved relevance over YouTube's default ordering, thereby enhancing the user experience.
\section{Dataset}\label{dataset}
\subsection{Data Collection}
To the best of our knowledge, no publicly available datasets exist for relevance and sentiment analysis of Bangla comments on YouTube movies and dramas. Therefore, we developed the 'CineXDrama' dataset using YouTube as the primary source for movie and drama comments, which were manually gathered and preprocessed to streamline the annotation.
We collected a total of 14,000 Bangla comments from YouTube. To build our dataset, we assembled potential comments from 36 different dramas and 25 movies. The YouTube Data API is used to retrieve user comments from the designated videos in a systematic way. More comments in this study came from dramas, with 7,772 comments, followed by movies with 6,228 comments. The data was organized in an Excel sheet with columns for the comments, their Relevance (relevant or irrelevant), and sentiment (positive or negative for relevant comments). Using this two-class Relevance detection structure, we annotated each comment as either relevant or irrelevant and further classified the relevant comments by sentiment. Overall, 56\% of the comments were deemed relevant, while 44\% were deemed irrelevant. Among the relevant comments, 55\% were positive, while 45\% were negative. The dataset is publicly available on 
\href{https://www.kaggle.com/datasets/pronaydebnath1/bangla-movie-drama-dataset}{Kaggle} \footnote{\textcolor{blue}{https://www.kaggle.com/datasets/pronaydebnath1/bangla-movie-drama-dataset}}.
\subsection{Data Preprocessing}
To reduce annotation workload and improve consistency, we applied a number of preprocessing filters to the collected comments, ensuring only validated comments were sent for manual annotation. We filtered out English comments, Banglish (mixed Bengali and English) comments, removed emojis from comments, discarded duplicate entries, along with sentences under three words, as they did not provide valuable information. These steps helped refine the dataset for further analysis.
\subsection{Annotation Guidelines}
Clear annotation guidelines are essential for maintaining quality and consistency in labeling. To annotate Bangla comments on movies or dramas, we first defined criteria to classify comments as relevant or irrelevant. After identifying relevant comments, we categorized them by sentiment, distinguishing between positive and negative sentiments (Figure \ref{fig:guideline}). Guidelines for annotators regarding relevance and sentiment are provided below:

\begin{figure}
  \centering
  \includegraphics[scale=0.7]{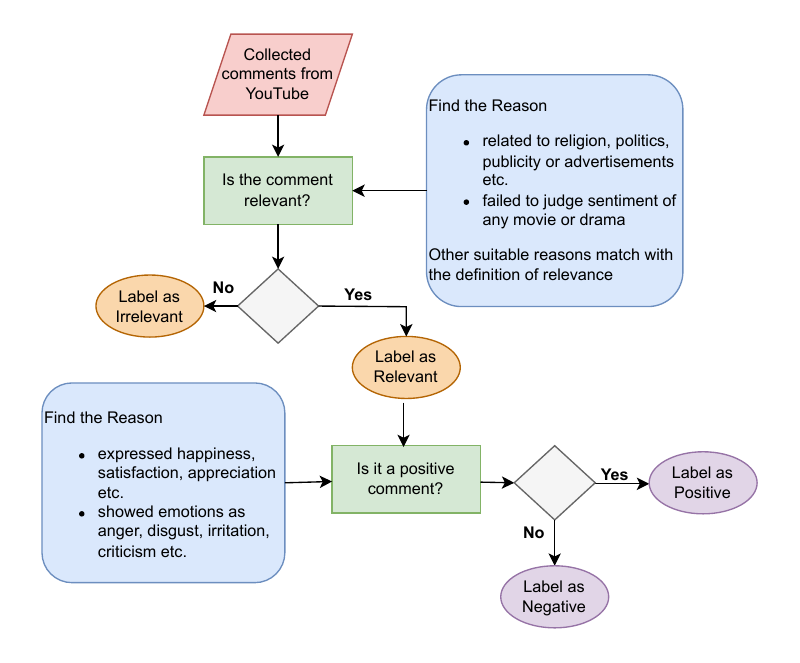}
  \caption{ Data annotation procedure along with pre-defined guidelines}
  \label{fig:guideline}
\end{figure}

\subsubsection{Relevance}Comments were marked as irrelevant if they discussed unrelated topics such as religion, politics, publicity, or advertisements, or if they lacked a clear sentiment judgment or connection to any drama or movie. If a comment didn’t indicate whether a movie or drama was perceived positively or negatively, it was also classified as irrelevant.
\subsubsection{Sentiment}After filtering out irrelevant comments, we categorized the relevant ones as either positive or negative. Negative comments showed emotions like anger, disgust, irritation, and criticism, while positive comments expressed happiness, satisfaction, or appreciation.

Here table \ref{tab:table for comment} shows sample comments from our dataset.
\begin{table}[http]
    \begin{center}
         \caption{Sample YouTube Comments}
          \label{tab:table for comment}
          \begin{tabular}{|>{\centering\arraybackslash}m{5.2cm}|
                           >{\centering\arraybackslash}m{1.1cm}|
                           >{\centering\arraybackslash}m{1.1cm}|}
            \hline
            \textbf{Comments} &
            \textbf{\centering Relevance} &
            \textbf{\centering Sentiment} \\
            \hline
            \textbf{{\bng Ek kthay AsadharN ichl naTkiT} (In one word, the drama was outstanding)} & Relevant  & Positive \\
            \hline
            \textbf{{\bng ipLj Aapnara Aamar cYaenliT sabs/kRa{I}b ker edn} (Please subscribe to my channel)} & Irrelevant & - \\
            \hline
            \textbf{{\bng Aaphran inesha ik Aibhny! bhaba Jay Asadharn} (What an acting by Afran Nisho! It's simply amazing)} & Relevant & Positive \\
            \hline
            \textbf{{\bng E{I} muibh Ta {I}slam iberadhii. {I}slam ek Ekhaen incu kra Heyech, ETa iThk Hyin} (This movie is anti-Islam. It has belittled Islam, which is not right)} & Relevant & Negative \\
            \hline
           \end{tabular}
\end{center}
\end{table}

\subsection{Dataset Statistics}
The dataset overview is summarized in table \ref{tab:table for stat}. This summary highlights both the relevance and sentiment distribution within the dataset.
\begin{table}[http]
    \begin{center}
         \caption{Dataset Statistics}
          \label{tab:table for stat}
          \begin{tabular}{|>{\centering\arraybackslash}m{2cm}|
                           >{\centering\arraybackslash}m{1.2cm}|
                           >{\centering\arraybackslash}m{1.1cm}|
                           >{\centering\arraybackslash}m{1cm}|
                           >{\centering\arraybackslash}m{1.1cm}|}
            \hline
            \textbf{Statistics} &
            \textbf{\centering Relevant} &
            \textbf{\centering Irrelevant} &
            \textbf{\centering Positive} &
            \textbf{\centering Negative} \\
            \hline
            \textbf{Total No. Data} & 7839 & 6161 & 4275 & 3564 \\
            \hline
            \textbf{Total word count} & 76959 & 57440 & 40642 & 36317 \\
            \hline
            \textbf{Total Unique Words} & 12953 & 12469 & 7407 & 8133 \\
            \hline
            \textbf{Maximum Comment Length} & 112 & 194 & 112 & 110 \\
            \hline
            \textbf{Average Number of Words} & 9.82 & 9.32 & 9.51 & 10.19 \\
            \hline
            
           \end{tabular}
\end{center}
\end{table}

\section{Methodology}
\begin{figure*}
  \centering
  \includegraphics[width=530pt, height=185pt]{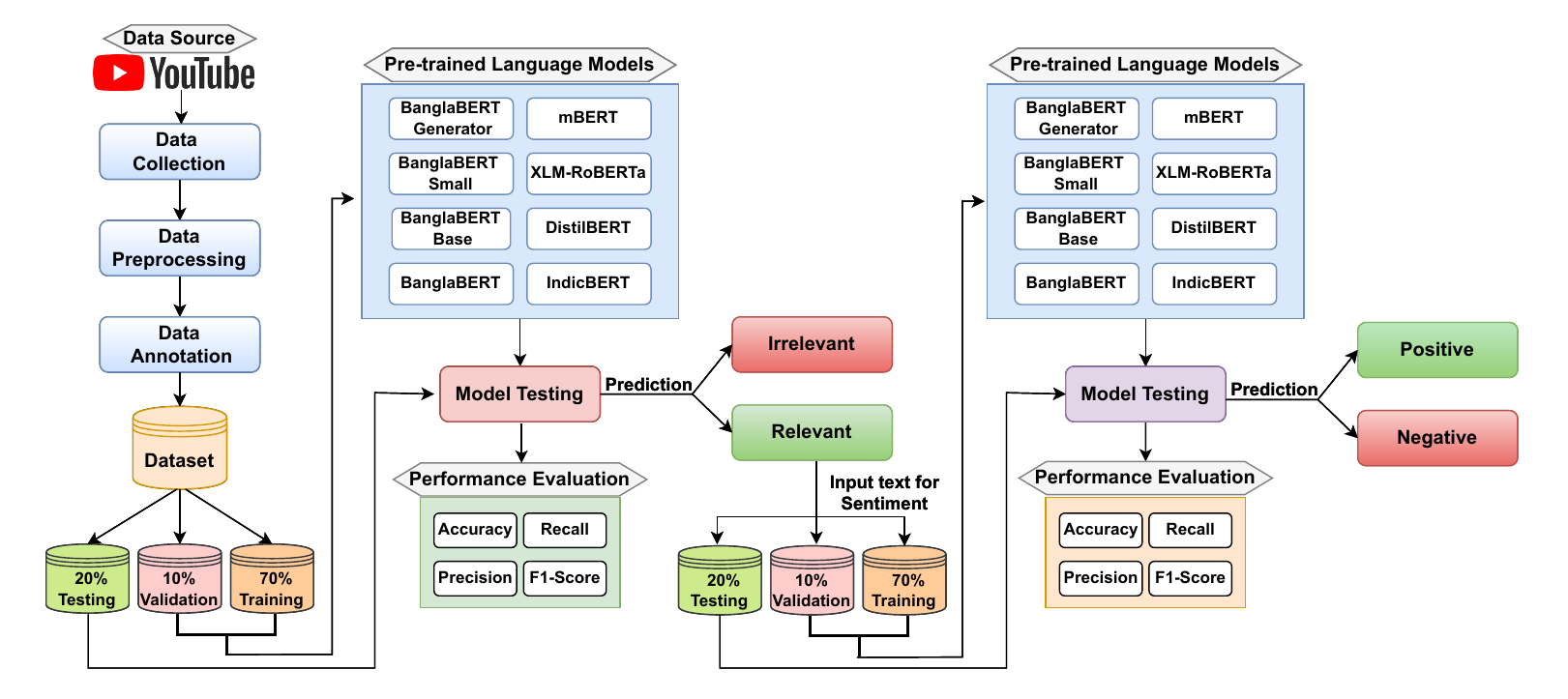}
  \caption{Suggested architecture of PLMs for relevance and sentiment analysis}
  \label{fig:SVM-framework}
  \vspace{-2ex}
\end{figure*}
\subsection{Dataset Development Pipeline}
The dataset development pipeline includes data collection, text preprocessing, data annotation, and data splitting. We gathered comments from Bangla drama-movies on YouTube and organized them in an Excel sheet. For preprocessing, we removed English and Banglish comments, eliminated duplicate sentences, discarded emojis etc. We then annotated the comments for relevance and sentiment according to our annotation guidelines (breifed in section \ref{dataset}). Finally, divided the dataset into training, testing, and validation sets.
\subsection{Classification using PLMs}
For both Relevance detection and sentiment analysis, we used eight pre-trained language models (PLMs). Using appropriate tokenizers for each pre-trained model, the text data was first tokenized. After that, we trained the models for a fixed number of epochs, updating the model parameters with the AdamW optimiser and measuring the classification errors with the cross-entropy loss function. The performance of the models was evaluated on a validation set at the end of each epoch, and learning progression and overfitting were monitored by tracking accuracy and loss metrics.
\subsection{Performance Evaluation}
We tested our model with the test set data and evaluated its performance using metrics such as accuracy \cite{accuracy}, precision \cite{preref1}, recall \cite{preref1}, and F1 score \cite{preref1}. Additionally, we generated confusion matrices and also interpreted the result using LIME for a clearer understanding.
\section{Experiments and Result Analysis}
\subsection{Experimental Setup} \label{setup}
For experimentation purpose, we utilized eight transformer models: BanglaBERT \cite{banglabert}, BanglaBERT Small \cite{banglabert}, Bangla BERT Base \cite{banglabertbase}, BanglaBERT Generator \cite{banglabert}, mBERT \cite{mBert}, XLM-RoBERTa \cite{xlmRoberta}, IndicBERT \cite{indicBert}, and DistilBERT \cite{distilBERT}. The experiments were conducted using the NVIDIA T4 GPUs with 16 GB of RAM on Google Colab and Kaggle. PyTorch (version 2.4.0) was used to develop and train the models.
\subsection{Hyperparameter Tuning}
In this experiment, the dataset was split into 70\% for training, 10\% for validation, and 20\% for testing. Also, we trained the models for 10 epochs using the Adam optimizer with a batch size of 16 and a learning rate of 2e-5.
\subsection{Results}
\subsubsection{Quantitative Analysis}
This part aims to illustrate the performance of relevance detection and sentiment analysis through numerical and visual representations, including tables, confusion matrices, and concise descriptions of the results.
\newline
\textbf{{Relevance Detection:}} Table \ref{tab:table_Relevance} demonstrates the superior performance of BanglaBERT compared to other transformer models for Relevance detection, achieving an accuracy of 83.99\%, a precision of 0.8442, a recall of 0.8310, and an F1 score of 0.8349. IndicBERT, on the other hand, had the lowest performance, with an accuracy of 77.71\%, a precision of 0.7750, a recall of 0.7704, and an F1 score of 0.7721. 
\begin{table}[http]
    \begin{center}
         \caption{Evaluation Table of Relevance Detection}
         \label{tab:table_Relevance}
         \begin{tabular}{|>{\centering\arraybackslash}m{2cm}|
                           >{\centering\arraybackslash}m{1.5cm}|
                           >{\centering\arraybackslash}m{1cm}|
                           >{\centering\arraybackslash}m{1cm}|
                           >{\centering\arraybackslash}m{1.1cm}|}
            \hline
            \textbf{Models} & \textbf{Accuracy(\%)} & \textbf{Precision} & \textbf{Recall} & \textbf{F1-score} \\       
            \hline
            \textbf{BanglaBERT (small)} & 80.89\% & 0.8200 & 0.7950 & 0.8000 \\
            \hline
            \textbf{BanglaBERT Generator} & 81.74\% & 0.8209 & 0.8078 & 0.8115 \\
            \hline
            \textbf{\textbf{BanglaBERT}} & \textbf{83.99\%} & \textbf{0.8442} & \textbf{0.8310} & \textbf{0.8349} \\
            \hline
            \textbf{mBert} & 78.63\% & 0.7840 & 0.7874 & 0.7847 \\
            \hline
            \textbf{XLM-RoBERTa} & 80.89\% & 0.8242 & 0.7933 & 0.7987 \\
            \hline
            \textbf{Bangla BERT Base} & 79.85\% & 0.7977 & 0.7915 & 0.7937 \\
            \hline
            \textbf{IndicBERT} & 77.71\% & 0.7750 & 0.7704 & 0.7721 \\
            \hline
            \textbf{DistilBERT} & 80.03\% & 0.7973 & 0.7992 & 0.7981 \\
            \hline
          \end{tabular}
    \end{center}
\end{table}
\begin{figure}
  \centering
  \includegraphics[scale=0.4]{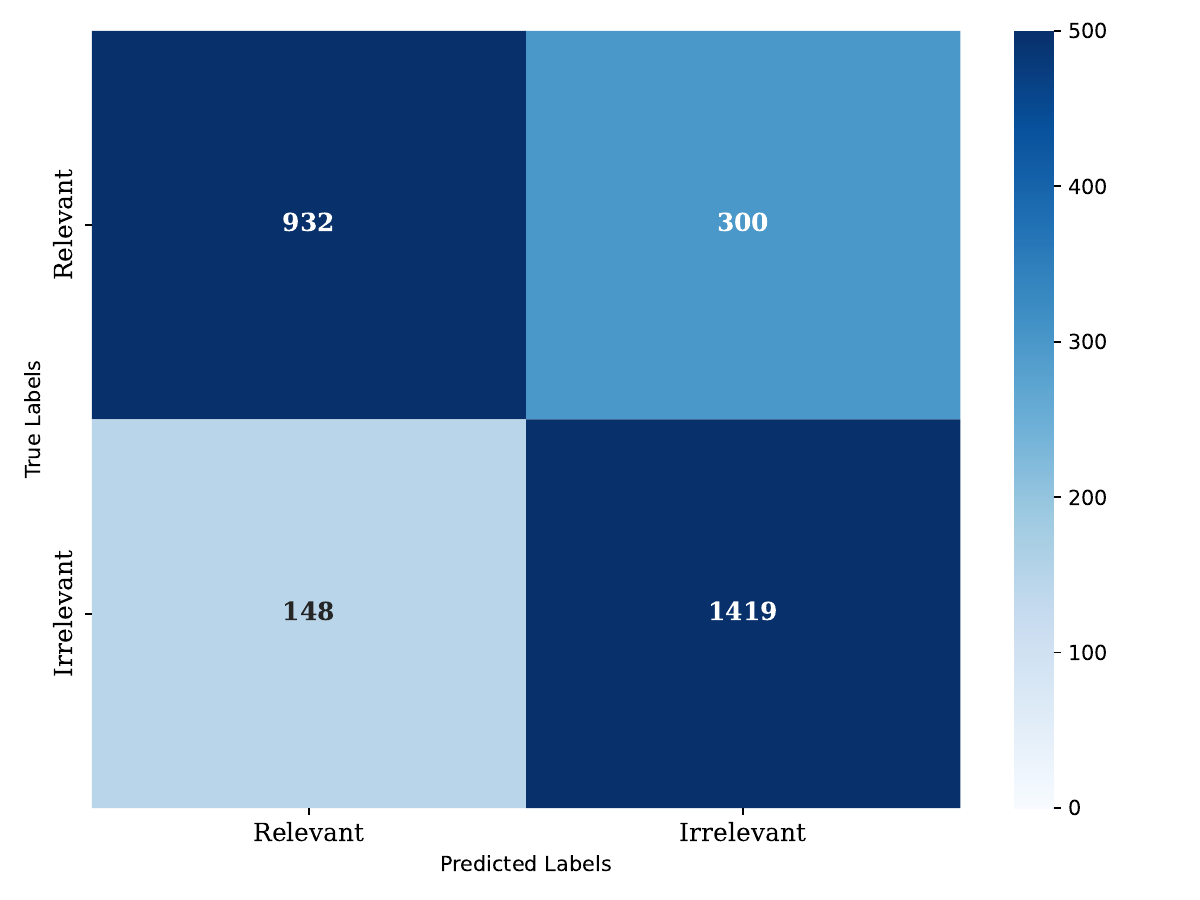}
  \caption{Confusion Matrix of BanglaBERT for Relevance Detection}
  \label{fig:rel_cm_banBERT}
\end{figure}
The results presented in Table \ref{tab:table_Relevance} are further supported by the confusion matrix in Figure \ref{fig:rel_cm_banBERT}. BanglaBERT demonstrates good accuracy, with its actual vs predicted values being closely aligned.
\newline
\textbf{{Sentiment Analysis:}}
Just as BanglaBERT performed the best in Relevance detection in Table \ref{tab:table_Relevance}, it also demonstrated superior performance in sentiment analysis with an accuracy of 93.30\%, a precision of 0.9338, a recall of 0.9311, and a F1-score of 0.9322, as we can see in Table \ref{tab:table for Sentiment}. While the other models also performed well comparatively, IndicBERT had the lowest performance, with an accuracy of 82.51\%, a precision of 0.8406, a recall of 0.8145 and a F1-score of 0.8185.
\begin{table}[http]
    \begin{center}
         \caption{Evaluation Table of Sentiment Analysis}
          \label{tab:table for Sentiment}
          \begin{tabular}{|>{\centering\arraybackslash}m{2cm}|
                           >{\centering\arraybackslash}m{1.5cm}|
                           >{\centering\arraybackslash}m{1cm}|
                           >{\centering\arraybackslash}m{1cm}|
                           >{\centering\arraybackslash}m{1.1cm}|}
            \hline
            \textbf{Models} &
            \textbf{\centering Accuracy(\%)} &
            \textbf{\centering Precision} &
            \textbf{\centering Recall} &
            \textbf{\centering F1-score} \\
            \hline
            \textbf{BanglaBERT (small)} & 90.11\% & 0.9022 & 0.8983 & 0.8998 \\
            \hline
            \textbf{BanglaBERT Generator} & 89.53\% & 0.8996 & 0.8905 & 0.8934 \\
            \hline
            \textbf{\textbf{BanglaBERT}} & \textbf{93.30}\% & \textbf{0.9338} & \textbf{0.9311} & \textbf{0.9322} \\
            \hline
            \textbf{mBert} & 89.78\% & 0.8970 & 0.9002 & 0.8975 \\
            \hline
            \textbf{XLM-RoBERTa} & 90.81\% & 0.9071 & 0.9100 & 0.9078 \\
            \hline
            \textbf{Bangla BERT Base} & 88.06\% & 0.8833 & 0.8764 & 0.8787 \\
            \hline
            \textbf{IndicBERT} & 82.51\% & 0.8406  & 0.8145 & 0.8185 \\
            \hline
            \textbf{DistilBERT} & 89.02\% & 0.8889 & 0.8913 & 0.8897 \\
            \hline
            
           \end{tabular}
\end{center}
\end{table}

\begin{figure}
  \centering
  \includegraphics[scale=0.4]{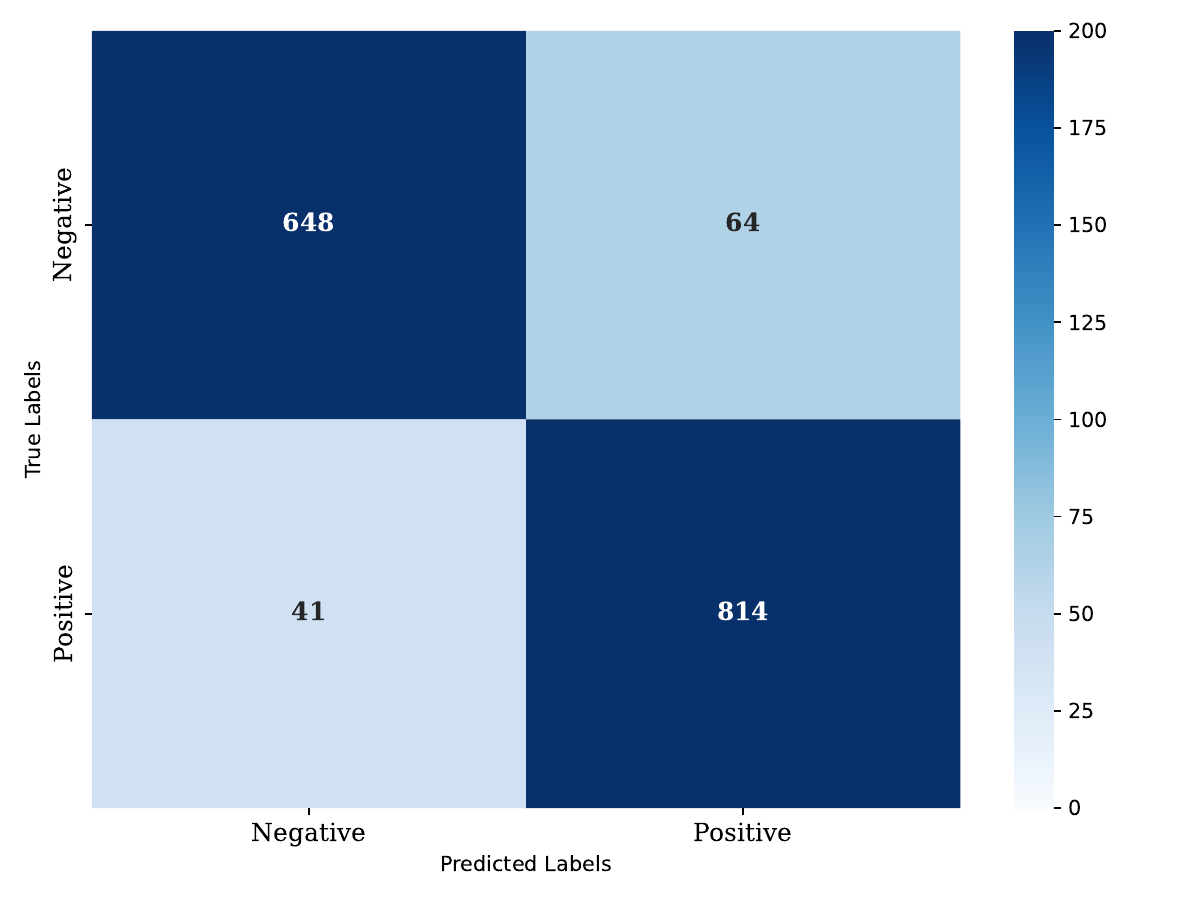}
  \caption{Confusion Matrix of BanglaBERT for Sentiment Analysis}
  \label{fig:senti_cm_banBERT}
\end{figure}
In Figure \ref{fig:senti_cm_banBERT}, we can see that BanglaBERT has performed well in predicting sentiment, as evident from the confusion matrix.
\subsubsection{Qualitative Analysis}
Among the eight transformers tested, four are specifically pretrained for the Bangla language (BanglaBERT, BanglaBERT base, BanglaBERT small, and BanglaBERT generator), while the other four are multilingual models (mBERT, XLM-RoBERTa, DistilBERT, and IndicBERT). BanglaBERT performed the best in both relevance detection and sentiment analysis tasks due to its extensive pretraining on a large amount of Bangla text, along with its higher number of parameters, which allow it to capture complex linguistic features specific to Bangla, providing it with a significant advantage for Bangla text classification. On the other hand, IndicBERT, which isn’t specifically trained for Bangla and has fewer parameters, showed the lowest performance, as expected. More visual result interpretation is displayed in subsection \ref{Lime} with LIME for better understanding of the results.
\subsubsection{XAI Explaination (Lime)\label{Lime}}
In order to provide insight into the features that most significantly affect the model’s decision-making process, we employed the Local Interpretable Model Agnostic Explanations (LIME) approach to analyze and visualize the predictions produced by our model. Considering two examples: one for a relevant comment that is correctly classified by BanglaBERT but misclassified by IndicBERT (for Relevance detection) and another for a negative comment that is correctly classified by BanglaBERT but misclassified by IndicBERT (for sentiment analysis). For comparison, we picked IndicBERT as the worst-performing model and BanglaBERT as the best-performing model.

\begin{figure}[!t]
    \centering
    \begin{minipage}[t]{0.48\textwidth}
        \centering
        \includegraphics[width=\textwidth]{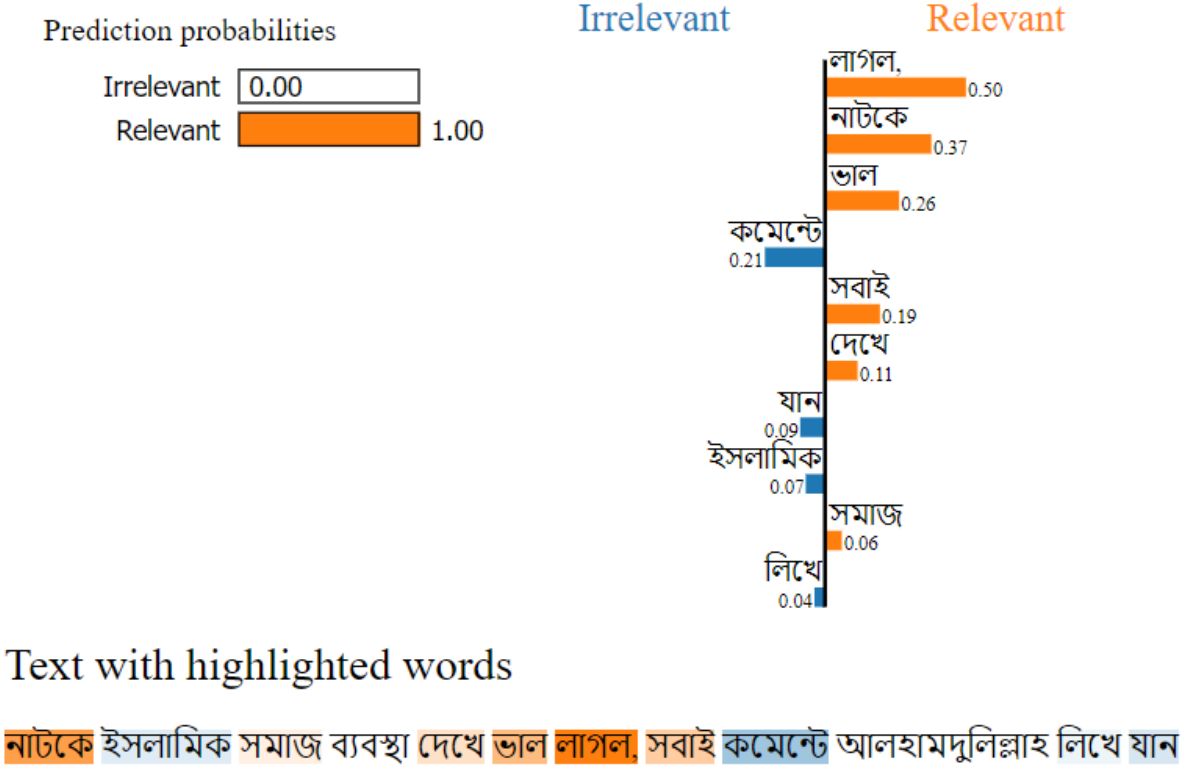}
        \caption{Relevant Comment Correctly Classified by BanglaBERT}
        \label{fig:image1}
    \end{minipage}
    \hfill
    \begin{minipage}[t]{0.48\textwidth}
        \centering
        \includegraphics[width=\textwidth]{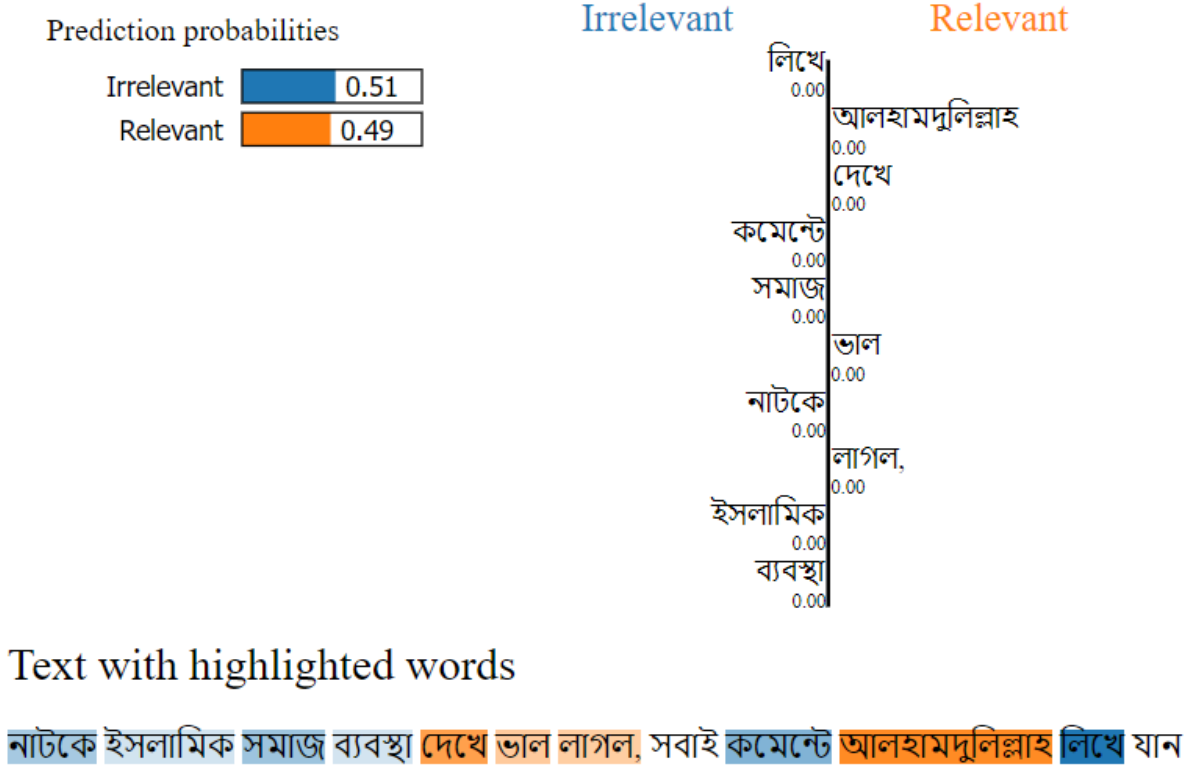}
        \caption{Relevant Comment Misclassified by IndicBERT}
        \label{fig:image2}
    \end{minipage}
\end{figure}

In Figure \ref{fig:image1}, we observe that the comment ‘{\bng naTek {I}slaimk smaj bYbs/tha edekh bhal lagl, sba{I} kemen/T AalHamduillLaH ilekh Jan}’ (It felt good to see the Islamic social system in the drama, everyone write Alhamdulillah in the comments) is correctly classified as relevant by BanglaBERT as the intensity of the words ‘{\bng naTek}’ (in the drama), ‘{\bng bhal}’(good), and ‘{\bng lagl}’ (felt) is quite significant and also all are contextually relevant to the drama. In contrast, IndicBERT focuses on the features ‘{\bng {I}slaimk}’ (Islamic), ‘{\bng smaj}’ (society), ‘{\bng kemen/T}’ (in the comments), and ‘{\bng ilekh}’ (write), which highlight a religious aspect and fail to capture the overall context of the comment, resulting in its misclassification as irrelevant, as seen in Figure \ref{fig:image2}.

\begin{figure}[!t]
    \centering
    \begin{minipage}[t]{0.48\textwidth}
        \centering
        \includegraphics[width=\textwidth]{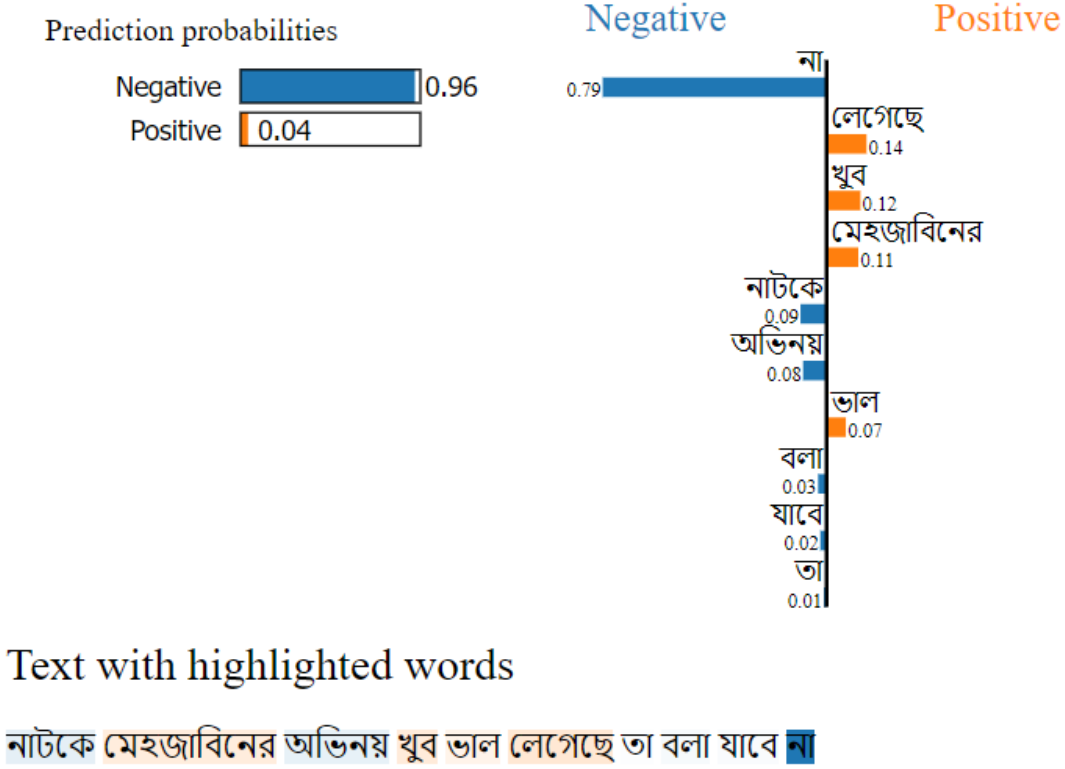}
        \caption{Negative Comment Correctly Classified by BanglaBERT}
        \label{fig:image3}
    \end{minipage}
    \hfill
    \begin{minipage}[t]{0.48\textwidth}
        \centering
        \includegraphics[width=\textwidth]{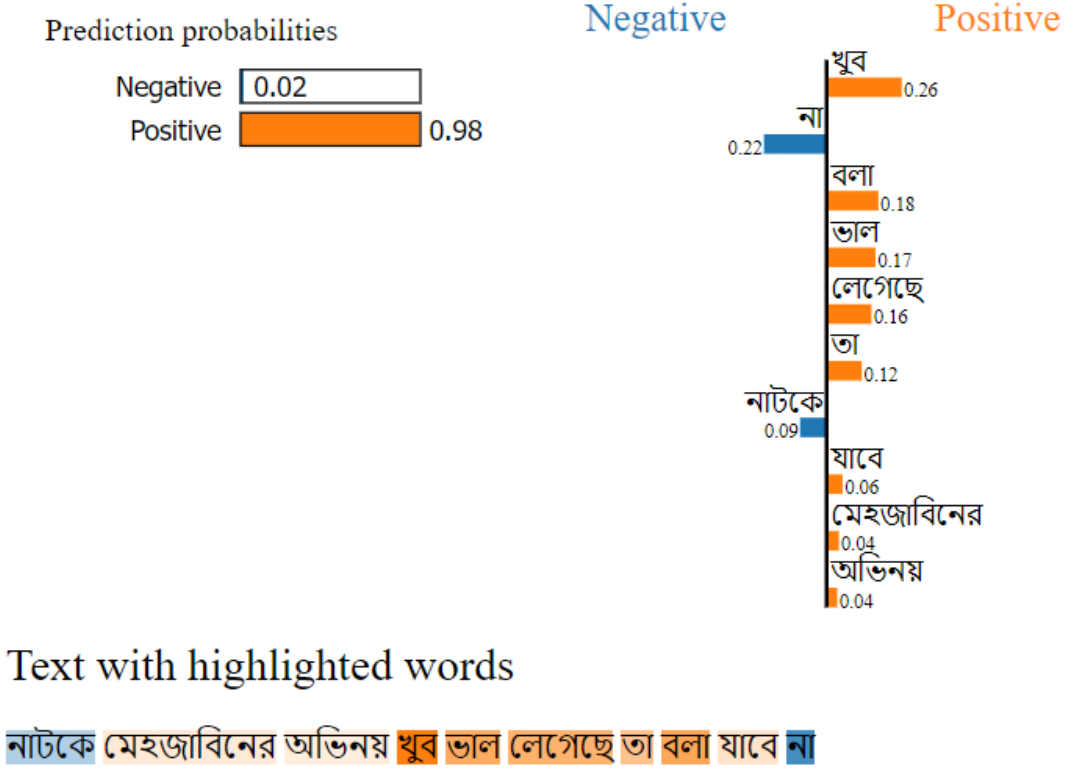}
        \caption{Negative Comment Misclassified by IndicBERT}
        \label{fig:image4}
    \end{minipage}
\end{figure}
‘ ’
In Figure \ref{fig:image3}, the comment ‘{\bng naTek emHjaibenr Aibhny khub bhal elegech ta bla Jaeb na}’ (It can not be said that Mehzabeen's acting in the drama was very good) is correctly classified as negative by BanglaBERT, whereas in Figure \ref{fig:image4}, IndicBERT misclassifies it. This is because IndicBERT primarily focuses on the words ‘{\bng khub}' (very), ‘{\bng bhal}' (good), and ‘{\bng elegech}' (felt), which are positive in nature, while giving less priority to the word ‘{\bng na}' (not) at the end, which negates the sentiment. This mix of positive and negative words causes confusion for IndicBERT, leading to the incorrect classification.

\section{Conclusion and Future Work}
This paper introduced a sentiment and relevance analysis system for Bangla YouTube comments on movies and dramas. To accomplish this, we built the ‘CineXDrama’ dataset, which contains 14,000 comments. Employing eight transformer models, we effectively classified sentiments and identified relevant content, while LIME provided interpretability to enhance prediction transparency. Our work fills a research gap in Bangla sentiment analysis, specifically for movie and drama reviews. Notably, BanglaBERT achieved the highest accuracy for both relevance and sentiment analysis because it effectively understands the unique language features and context of Bangla. For future work, we aim to expand the sentiment categories to include emotions such as sad, angry, and happy, etc. Additionally, expanding the dataset to include comments from other forms of media, like music videos, and vlogs, will enrich our analysis beyond just dramas and movies. This broader approach will provide deeper insights and more meaningful feedback for the Bangladeshi entertainment industry.

\bibliographystyle{IEEEtran}
\bibliography{pronay}

\end{document}